\documentclass{article}

\PassOptionsToPackage{numbers,compress,sort}{natbib}
\usepackage[preprint]{neurips_2025}

\usepackage[utf8]{inputenc} 
\usepackage[T1]{fontenc}    
\usepackage{url}            
\usepackage{booktabs}       
\usepackage{amsfonts}       
\usepackage{nicefrac}       
\usepackage{microtype}      
\usepackage{xcolor}         
\usepackage{graphicx}
\usepackage{amsmath}
\usepackage{amssymb}
\usepackage{pifont}
\usepackage{wrapfig}
\definecolor{lightgray}{gray}{0.85} 
\newcommand{\xmark}{\textcolor{lightgray}{\ding{55}}} 
\newcommand{\cmark}{\checkmark} 
\usepackage{enumitem}
\usepackage{tcolorbox}
\definecolor{blue}{rgb}{0.21,0.49,0.74}
\definecolor{red}{rgb}{0.8, 0.2, 0.2}
\definecolor{green}{rgb}{0, 0.5, 0}
\definecolor{yellow}{RGB}{218, 160, 109}
\usepackage[pagebackref=true,breaklinks=true,colorlinks=true,citecolor=blue,linkcolor=red,bookmarks=false]{hyperref}

\title{DRAMA-X: A Fine-grained Intent Prediction and Risk Reasoning Benchmark For Driving}

\author{%
   \bf Mihir Godbole, Xiangbo Gao, Zhengzhong Tu\thanks{Corresponding Author: Zhengzhong Tu (\texttt{tzz@tamu.edu})}\\[2pt]
  Texas A\&M University
}

\begin{document}

\maketitle

\begin{abstract}
\label{abs}
Understanding the short-term motion of vulnerable road users (VRUs) like pedestrians and cyclists is critical for safe autonomous driving, especially in urban scenarios with ambiguous or high-risk behaviors. While vision-language models (VLMs) have enabled open-vocabulary perception, their utility for fine-grained intent reasoning remains underexplored. Notably, no existing benchmark evaluates multi-class intent prediction in safety-critical situations, To address this gap, we introduce \textbf{DRAMA-X}, a fine-grained benchmark constructed from the DRAMA dataset via an automated annotation pipeline. DRAMA-X contains 5,686 accident-prone frames labeled with object bounding boxes, a nine-class directional intent taxonomy, binary risk scores, expert-generated action suggestions for the ego vehicle, and descriptive motion summaries. These annotations enable a structured evaluation of four interrelated tasks central to autonomous decision-making: object detection, intent prediction, risk assessment, and action suggestion.
As a reference baseline, we propose SGG-Intent, a lightweight, training-free framework that mirrors the ego vehicle’s reasoning pipeline. It sequentially generates a scene graph from visual input using VLM-backed detectors, infers intent, assesses risk, and recommends an action using a compositional reasoning stage powered by a large language model. We evaluate a range of recent VLMs, comparing performance across all four DRAMA-X tasks. Our experiments demonstrate that scene-graph-based reasoning enhances intent prediction and risk assessment, especially when contextual cues are explicitly modeled. Our code and dataset is available at: \url{https://github.com/taco-group/DRAMA-X}.

\end{abstract}

\section{Introduction}
\label{intro}

Autonomous driving systems are structured around four sequential stages—perception, prediction, planning, and behavior execution—each supported by specialized sensor suites and data-driven models. In perception, multi-modal inputs such as multiple cameras, LIDAR and radar, are fused to detect and localize agents. Recent work has enhanced this stage with vision-language models (VLMs) \cite{najibi2023unsupervised, yao2023detclipv2, tian2024drivevlm, gao2025langcoop, wang2025generative, xing2024autotrust, luo2025v2x, gao2025automated} for open-vocabulary detection and richer semantic scene understanding. End-to-end  autonomous driving (E2E-AD) systems have also have been developed with the help of VLM capabilites \cite{zheng2024simplellm4ad, xu2024drivegpt4, you2024v2x, jiang2024senna, Mao2023GPTDriver, Tami2024MLLMEvents, xing2025openemma}. Despite these advances, real-world deployments still exhibit elevated accident rates in complex urban scenarios, particularly involving vulnerable road users. Analyses of autonomous vehicle incident reports reveal that pedestrians and cyclists contribute disproportionately to collision risk \cite{kutela2022mining, abdel2024matched}. Because an Autonomous Vehicle’s planning and behavior modules depend on accurate predictions of the immediate goals of other agents, improving intent prediction for pedestrians and cyclists is essential for enabling timely, safety-critical maneuvers.  

Intent prediction for vulnerable road users (VRUs) is a crucial component of urban autonomous driving~\cite{li2025simulating}, as it informs timely and safety-critical planning decisions. Early research framed this task as a binary problem—typically predicting whether a pedestrian would cross the street—enabled by datasets such as JAAD \cite{Rasouli2017JAAD} and PIE \cite{Rasouli2019PIE}. Over time, models evolved to incorporate richer cues like pose dynamics, motion trajectories, and social interactions, leading to increasingly complex architectures including attention-based, graph-based, and transformer-style networks \cite{sadeghian2019sophie, alahi2016social, zhou2021social, ling2024pedast}. While these advances mark significant progress, existing approaches remain limited in three key ways: (1) they primarily focus on coarse-grained binary or ternary intent labels; (2) they concentrate on pedestrians while largely ignoring cyclists, whose behaviors often differ but are equally critical; and (3) they emphasize routine crossing scenarios rather than the high-risk or ambiguous events—such as jaywalking, sudden veering, or mid-road hesitation, where intent recognition is most safety-relevant. These limitations motivate the need for benchmarks that encompass multi-class, multi-actor intent labels in truly hazardous driving scenes.

To address these limitations, we introduce \textbf{DRAMA-X}, a benchmark derived from the DRAMA dataset \cite{malla2023drama}, comprising 5686 curated urban driving scenarios featuring both pedestrians and cyclists in risky situations. Each scenario is densely annotated with object bounding boxes for all dynamic agents; a nine-class directional intent taxonomy for each VRU such as "moves to the right", "moves to the left", "stationary"; frame-level binary risk scores based on the perceived danger; and expert-recommended ego-vehicle actions (e.g., “yield,” “brake and hold,” “proceed with caution”). We utilize object motions from ground truth scene videos to ensure the validity of generated annotations. We define four interrelated evaluation tasks: object detection, intent classification, risk assessment, and action suggestion—and propose \textbf{SGG-Intent}, a modular baseline pipeline. SGG-Intent first generates a structured scene graph capturing spatial and semantic relationships (leveraging a VLM-backed detector and a graph neural network), then sequentially infers VRU intents, assesses overall scene risk, and produces action recommendations via a lightweight LLM stage. This sequential design mirrors human-like reasoning by conditioning each downstream task on contextual cues extracted in earlier stages. We utilize the open-vocabulary detection capability of VLMs to detect objects and model relationships to output a scene graph. We conducted experiments to evaluate the efficacy of different state-of-the-art VLMs \cite{bai2025qwen2, OpenAI2023GPT4, deitke2024molmo, li2024llava} and reported the results on our DRAMA-X benchmark. Our key contributions are summarized as follows:
\begin{itemize}[leftmargin=1em, itemsep=1pt, parsep=1pt, topsep=0pt, partopsep=0pt]
  \item We introduce \textbf{DRAMA-X}, the first large-scale benchmark to evaluate fine-grained intents for pedestrians and cyclists in high-risk scenarios. Our benchmark is built upon ground truth object motion trajectories from the original DRAMA dataset.
  \item To construct DRAMA-X, we design an \textbf{automatic annotation pipeline} that identifies and generates fine-grained intents for critical objects in the scene. 
  \item We design a novel intent prediction framework called \textbf{SGG-Intent}, which combines scene graph generation, VLM perception modules, and sequential reasoning for intent and risk. This framework exhibits improved scene understanding and reasoning over direct VLM querying. 
  \item We established a \textbf{benchmark} and provided empirical validations on DRAMA-X with four sequential tasks from perception (detection) to recommended driving actions (intent, risk, action). Our experiments evaluate the performance of VLMs on each task, with and without the proposed baseline framework. Additionally, we also show that the order of execution of these tasks affects the performance of downstream tasks.
\end{itemize}

\section{Related Work}
\label{rel_works}
\paragraph{Intent Prediction} Early pedestrian intention research focused on binary crossing prediction using dynamics and context cues \cite{Kooij2014Context,Schneemann2016Context}. Benchmarks like JAAD and PIE advanced data-driven approaches: JAAD provided 346 video clips with pedestrian behavior annotations \cite{Rasouli2017JAAD}, while PIE extended this with over six hours of driving footage, achieving $\sim$79\% crossing prediction accuracy using LSTM models \cite{Rasouli2019PIE}. Subsequent works incorporated multi-modal cues \cite{Fang2018, gao2024mambast} or Transformer architectures to capture temporal context \cite{Kotseruba2021Review}. However, these datasets address only coarse binary outcomes without distinguishing directional movement patterns or cyclist behavior, and they don't connect intentions to risk assessment. Our DRAMA-X dataset advances this research by providing fine-grained directional intention labels (seven classes) for both pedestrians and cyclists, while linking intent to explicit risk assessment. As noted in a recent survey \cite{Kotseruba2021Review}, DRAMA-X fulfills the need for richer annotations and sequential decision evaluation in this domain.

\paragraph{Autonomous Driving Datasets} Numerous driving benchmarks exist primarily for perception tasks\cite{gao2025airv2x}. KITTI provided camera/LiDAR data for 3D detection and tracking \cite{Geiger2013KITTI}, nuScenes added 1000 scenes with multi-sensor annotations for 23 object classes \cite{Caesar2020nuScenes}, and Waymo Open Dataset scaled to 1150 segments with high-resolution sensor streams \cite{Sun2020Waymo}. Argoverse introduced HD maps for enhanced tracking \cite{Chang2019Argoverse}, while BDD100K supplied 100k videos with diverse annotations \cite{Yu2020BDD100K}. Motion-forecasting datasets like Waymo Open Motion and Argoverse release agent trajectories but treat them as coordinate sequences without categorical intent reasoning \cite{ettinger2021large,Chang2019Argoverse}.Recent efforts incorporate higher-level reasoning: DriveLM uses graph-structured visual QA on nuScenes scenes \cite{Sima2024DriveLM}, WOMD-Reasoning adds free-form Q\&A pairs about interactions \cite{Li2024WOMD-Reasoning}, and BDD-X, BDD-OIA, and Reason2Drive augment driving videos with textual explanations \cite{Kim2018BDD-X,Xu2020OIA,Nie2024Reason2Drive}. These rely on free-form QA rather than standardized tasks. DRAMA-X bridges low-level trajectory datasets and language-guided benchmarks by providing structured intent and risk annotations with explicit evaluation metrics for detection, intent prediction, risk assessment, and action suggestion.

\paragraph{Scene Graph Generation} Conventional SGG methods typically separate object detection from relation classification, training on Visual Genome for fixed predicates \cite{Krishna2017VG, Xu2017SceneGraph, Zellers2018NeuralMotifs}. Prompting-only VLM approaches leverage large multimodal models: Mitra et al. use Compositional Chain-of-Thought prompting \cite{Mitra2023CCoT}, PRISM-0 bootstraps VLM captions into predicate-rich prompts \cite{Elskhawy2025PRISM}, and SG-Nav constructs hierarchical 3D scene graphs for object navigation \cite{Yin2024SGNav}.Training-based open-vocabulary frameworks integrate VLM pretraining: From Pixels to Graphs generates scene graphs via image-to-text pipelines \cite{Li2024PixelsToGraphs}, OvSGTR aligns visual regions with text embeddings \cite{Chen2024OvSGTR}, and EGTR extracts relation graphs from DETR's self-attention outputs \cite{Im2024EGTR}. Other works extend prompting to hierarchical settings \cite{Liu2024RAHP} and video domains \cite{Wang2024VidVRD}.





\section{DRAMA-X}
\label{sec:benchmark}

The DRAMA-X benchmark is derived from the DRAMA driving-risk dataset, originally containing 17,785 annotated urban traffic scenarios in Tokyo, Japan \cite{malla2023drama}. More details about this dataset can be found in Appendix.
The data curation and collection pipeline used to derive relevant samples from DRAMA dataset is detailed in \S \ref{sec:collection}. We introduce the automated data annotation pipeline in \S\ref{sec:annotation}. Finally, we present dataset statistics and examples in \S\ref{sec:stats}.

\subsection{Data Collection and Curation}
\label{sec:collection}
\paragraph{Collection.} We surveyed existing driving datasets to identify one suitable for intent-focused benchmarking with: (1) raw video data for temporal intent annotations, (2) abnormal/hazardous scenarios, (3) critical pedestrian and cyclist instances, and (4) intent annotation support. As shown in Table~\ref{dataset_comparison}, we selected the DRAMA dataset \cite{malla2023drama} for its frequency of annotated risky scenarios and partial intent labels. Alternative datasets like DADA-2020 \cite{fang2019dada}, DeepAccident \cite{wang2024deepaccident}, and CarCrash \cite{bao2020uncertainty} contain accident footage but lack sufficient pedestrian/cyclist interactions for quality intent annotation.
\paragraph{Curation.}
\begin{figure}
\centering
\includegraphics[width=\linewidth]{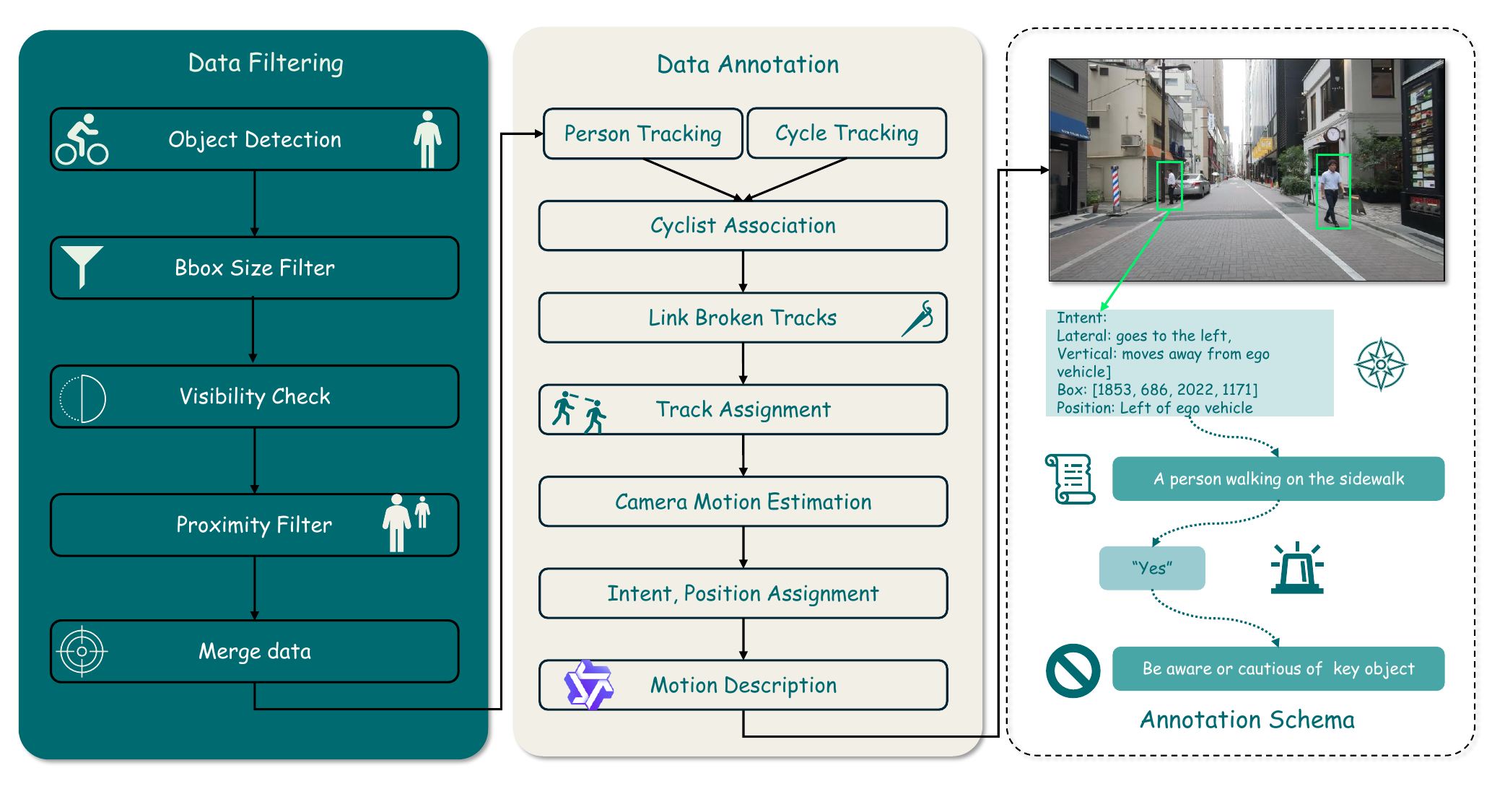}
\caption{Filtering and data annotation pipeline used to identify relevant scenarios with key VRU actors.}
\label{fig:filtering_pipeline}
\end{figure}
We developed a filtering pipeline (Figure~\ref{fig:filtering_pipeline}) to isolate scenes where pedestrians or cyclists represent potential risk. YOLOv8 \cite{ultralytics2023yolov8} was used to detect 'person' and 'cycle' instances, with a custom association function combining these into 'cyclist' labels. Spatial constraints through 'Bbox size filter' and 'Visibility check' ensured object salience. These detections were merged with existing DRAMA annotations to recover missed instances. The final DRAMA-X dataset preserves relevant fields (Risk, Suggestions) and stores each sample as a JSON object containing image URLs, object details (bounding boxes, intent, relative position), suggested actions, risk scores, and textual motion descriptions. Further details appear in the Appendix.

\begin{table}
  \caption{Comparison of datasets based on various characteristics, including object intent, video availability, and risk score.}
  \label{dataset_comparison}
  \centering
  \small
  \setlength{\tabcolsep}{5pt}
  \begin{tabular}{lccccccc}
    \toprule
    \textbf{Dataset}  & \textbf{Year} & \textbf{Pedestrians} & \textbf{Cyclists} & \textbf{Object Intent} & \textbf{Image} & \textbf{Video} & \textbf{Risk Score} \\
    \midrule
    WOMD            & 2021 & \cmark & \cmark & \cmark & \xmark & \cmark & \xmark \\
    WOMD-Reasoning          & 2024 & \cmark & \cmark & \cmark & \xmark & \cmark & \xmark \\
    DriveLM            & 2024 & \cmark & \cmark & Indirect & \cmark & \xmark & \xmark \\
    DHPR     & 2024 & \cmark & \cmark & Partial  & \cmark & \xmark & \cmark \\
    DADA-2020             & 2020 & \cmark & \xmark & \xmark   & \cmark & \cmark & \xmark \\
    DeepAccident   & 2023 & \cmark & \cmark & \xmark   & \cmark & \cmark & \cmark \\
    CarCrash (CCD)   & 2020 & \cmark & \cmark & \xmark   & \cmark & \cmark & \cmark \\
    DRAMA               & 2022 & \cmark & \cmark & Partial  & \cmark & \cmark & \cmark \\
    \midrule
    \textbf{DRAMA-X (ours)}                   & 2025 & \cmark & \cmark & \cmark   & \cmark & \cmark & \cmark \\
    \bottomrule
  \end{tabular}
\end{table}

\subsection{Annotation Pipeline}
\label{sec:annotation}

We design a multi-stage annotation pipeline to generate fine-grained object-level intent annotations in risky driving scenarios. This pipeline operates on video clips corresponding to the data samples filtered in the first stage and includes tracking, motion analysis, and natural language description generation. An overview of the process is shown in Figure~\ref{fig:filtering_pipeline}. Implementation details and parameters for the individual steps are provided in the Appendix.

\paragraph{Object Tracking.}
We employ a hybrid object tracking module to obtain dense object trajectories for vulnerable road users (VRUs). Person tracking is performed using YOLOv8 \cite{ultralytics2023yolov8} + BoT-SORT \cite{aharon2022bot}, while cyclist detection and tracking is handled using a Faster R-CNN model with a ResNet-50-FPN backbone \cite{ren2016faster} +  BoT-SORT. We post-process these detections to obtain clean trajectories for the objects in the scene.

Fragmenting is a common problem in object tracking. To mitigate this problem, we define a spatiotemporal affinity score based on predicted displacement, frame gap, and motion model confidence. Let $d_{\text{spatial}}$ be the Euclidean distance between the predicted end of track $T_i$ and the start of $T_j$, and $\Delta t$ be the temporal gap. The linking score $S_{i,j}$ is computed as:
\begin{equation*}
S_{i,j} = \left(1 - \frac{d_{\text{spatial}}}{D_{\max}}\right) \cdot w_s + \left(1 - \frac{\Delta t}{T_{\max}}\right) \cdot w_t \quad\quad  \hat{S}_{i,j} = S_{i,j} \cdot (0.5 + 0.5\alpha)
\end{equation*}
where $\alpha \in [0,1]$ is the trajectory prediction confidence, and $w_s$, $w_t$ are weights for spatial and temporal terms. $D_{\max}$ and $T_{\max}$ are adaptive thresholds based on track length and frame gap. A link is accepted if: $\hat{S}_{i,j} > \theta_{\text{link}}$.

\paragraph{Object Matching.}
Tracked objects are aligned with the filtered object set using bounding box overlap and global optimization. We first remove duplicate annotations, then apply the Hungarian algorithm to assign detected tracks to their corresponding filtered annotations, ensuring consistency in identity and motion context. To match filtered ground-truth boxes $\mathcal{B} = \{B_1, \dots, B_m\}$ with active tracked objects $\mathcal{T} = \{T_1, \dots, T_n\}$, we define a cost matrix $C \in \mathbb{R}^{n \times m}$ based on inverse IoU to find matches.

\paragraph{Camera Motion Estimation.}

To annotate ground relative intents, we decouple ego-motion from object motion by estimating the camera displacement between consecutive frames using median optical flow in regions adjacent to the object. Let $I_t$ and $I_{t+1}$ denote two consecutive video frames, and let $\mathcal{R}_t$ be a rectangular region around the object in $I_t$. The optical flow in the selected region is then averaged to estimate camera displacement as shown in Equation \ref{eq:opt-flow}, where $\mathbf{F}_t$ is the dense optical flow using the Farneback method. Object motion relative to the road is computed by subtracting this displacement from the object’s tracked motion. 
\begin{equation*}
\label{eq:opt-flow}
\Delta \mathbf{c}_t = \left( \frac{1}{|\mathcal{R}_t|} \sum_{(x,y) \in \mathcal{R}_t} \mathbf{F}_t(x, y) \right) = (dx_t^{\text{cam}}, dy_t^{\text{cam}})
\end{equation*}


\paragraph{Intent Inference.}
We define intent as immediate short-term motion without temporal dependency, distinct from trajectory prediction. DRAMA's short-duration videos are ideal for our annotation schema. For each tracked object, we compute displacement vectors across multiple temporal windows, adjusting for estimated camera motion to obtain road-relative displacement. Lateral intent is determined by the sign and magnitude of horizontal displacement, while vertical intent indicates whether the object moves toward or away from the ego vehicle in the image plane. The possible labels are:
\begin{tcolorbox}[colback=gray!5!white, colframe=gray!75!black, before skip=1mm, after skip=1mm, boxsep=0.0cm, middle=0.0cm, top=0.1cm, bottom=0.1cm]
    \textbf{Lateral intent values:} [\texttt{stationary},\ \ \texttt{goes to the left},\ \ \texttt{goes to the right}]
    
    \textbf{Vertical intent values:} [\texttt{stationary},\ \ \texttt{moves towards ego vehicle},\ \ \texttt{moves away from ego vehicle}]
\end{tcolorbox}
Additionally, the object's final $x$-coordinate is used to classify its relative position as \emph{Left}, \emph{Right}, or \emph{Front} of the ego vehicle. 

\paragraph{Language Annotation with Qwen2.5-VL.}
To generate interpretable annotations, we use Qwen2.5-VL to convert each object’s motion trajectory and intent metadata into a natural language description. This includes the object’s spatial behavior, inferred intent, and its relevance to the risk scene (e.g., “A cyclist is approaching the intersection from the right, likely to cross.”). This step supports downstream explainability and reasoning tasks.


\subsection{Dataset Statistics}

\label{sec:stats}

\begin{figure}[h]
\centering
\includegraphics[width=\linewidth]{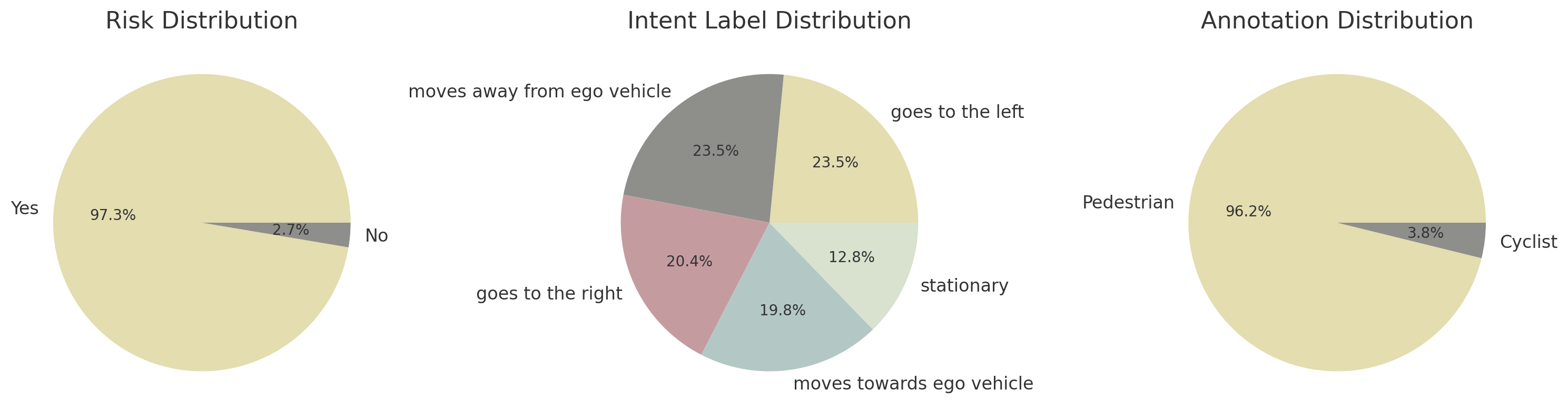}
\caption{Distribution characteristics of the DRAMA-X dataset. \textbf{Left:} Annotation distribution between pedestrians (96\%) and cyclists (4\%), reflecting typical urban road user composition. \textbf{Center:} Risk assessment distribution, with 97\% of scenarios containing potential risk situations that autonomous systems must address. \textbf{Right:} Intent label distribution across five movement categories, showing the number of instances for each directional intent (3,447 leftward, 3,454 away from ego vehicle, 3,000 rightward, 2,911 toward ego vehicle, and 1,874 stationary), providing balanced coverage of possible movement patterns.}
\label{fig:data_stats}
\end{figure}

The DRAMA-X benchmark comprises a diverse collection of 5,686 annotated video clips spanning approximately 54,892 frames. The dataset captures a comprehensive range of vulnerable road user scenarios, with 5,638 frames containing pedestrians and 361 frames featuring cyclists. We meticulously annotated 9,606 distinct object instances (9,237 pedestrians and 369 cyclists), ensuring thorough coverage of various interaction scenarios. Figure~\ref{fig:data_stats} illustrates the distribution characteristics of our dataset.

\section{SGG-Intent Framework} 
\label{sec:framework}
\begin{figure*}
\centering
\includegraphics[width=\linewidth]{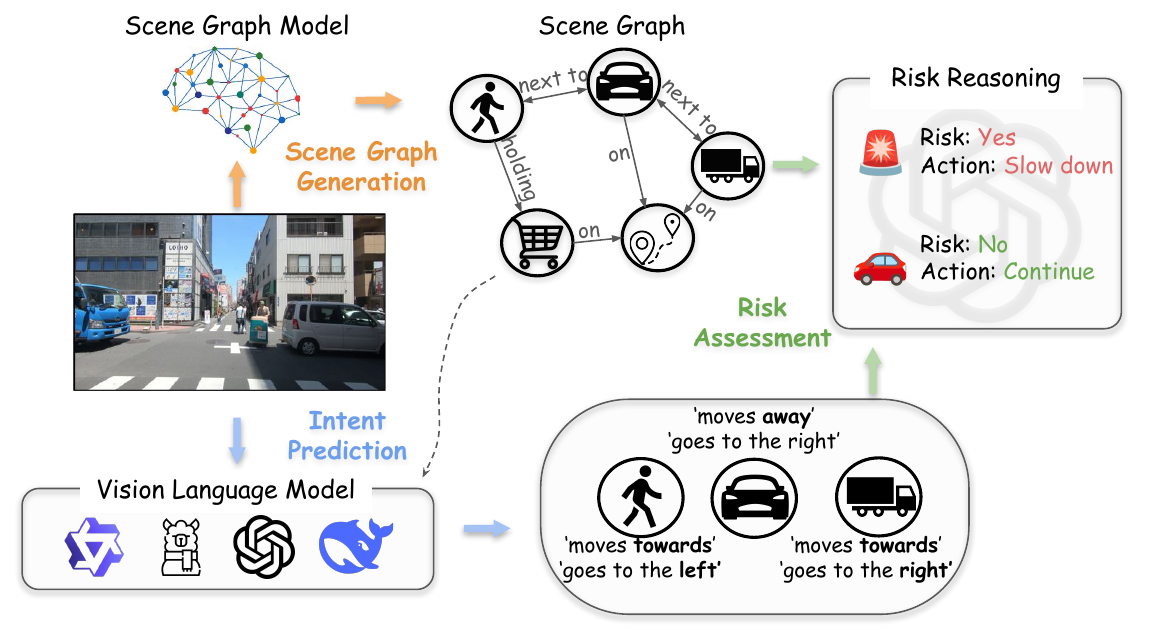}
\caption{Overview of the novel intent prediction \textbf{SGG-Intent} framework, for evaluation on DRAMA-X. Given an input image, a vision-language model (VLM) generates a scene graph capturing objects, attributes, and relationships. A separate intent prompt predicts directional motion intents for each dynamic object. These structured representations are passed to a risk reasoning LLM that assesses scene-level risk and suggests ego-vehicle actions (e.g., slow down, continue). The pipeline reflects the sequential decision-making process in autonomous driving.}

\label{sgg_framework}
\end{figure*}

To establish a baseline for this novel benchmark, we propose SGG-Intent, a training-free framework that leverages scene graph generation as a prior for intent prediction, as illustrated in Figure~\ref{sgg_framework}. Building upon recent advances in prompt engineering for single-pass scene graph generation \cite{mitra2024compositional}, we develop a specialized prompt structure and scene graph formalism tailored specifically for driving scenarios. For the final reasoning stages, we employ \texttt{GPT-4o} to generate risk assessments and action suggestions. Notably, our framework's modular design allows the Scene Graph Generation (SGG) component to be implemented using various approaches, from directly prompting Vision-Language Models to utilizing pre-trained scene graph generation models. When VLMs are incorporated into the pipeline, our specialized scene graph prompt significantly enhances their spatial reasoning capabilities and provides structured representations for downstream tasks.

\paragraph{Driving Scene Graph Generation}
Our framework utilizes a specialized scene graph representation optimized for driving scenarios. The scene graph captures spatial relationships between road agents (e.g., next to'', on''), as well as object attributes including aesthetic and physical orientation description. The graph structure enables efficient reasoning about potential interactions between vehicles, pedestrians, and other traffic elements, forming a foundation for accurate intent prediction and risk assessment. We require the model to output the scene graph in the following structured format:
This structured representation facilitates systematic spatial reasoning and provides a clear interface between the vision components and downstream reasoning modules.

\paragraph{Intent Prediction}
Building upon the generated scene graph, our framework employs a systematic approach to intent prediction. By analyzing the spatial and relational patterns captured in the scene graph, we identify potential movement trajectories and interaction intentions of various road agents. The intent prediction module focuses particularly on dynamic predicates like "moves towards" and "goes to the right" to forecast likely agent behaviors. The model is required to generate dual intent labels for every object in the scene graph. 

\paragraph{Risk Assessment and Action Suggestion}
The final stage of our SGG-Intent framework leverages the predicted intents to perform comprehensive risk assessment. Using the structured scene representation and predicted agent intentions, we employ \texttt{GPT-4o} to reason about potential hazards and safety implications. The system evaluates collision risks, traffic rule violations, and other safety concerns to generate a binary risk classification (Risk: Yes/No). For scenarios classified as risky, the framework recommends appropriate defensive driving actions (e.g., "Slow down"). This risk-aware action recommendation completes the perception-to-decision pipeline, demonstrating the practical utility of our scene graph-based approach.
\section{Experiments}
\label{sec:experiments}

\subsection{Experimental Setup}
To evaluate the proposed DRAMA-X benchmark on different methods including VLMs, we conduct experiments. We select some of the best performing VLMs on the NaturalBench benchmark for compositional understanding \cite{li2024naturalbench} including \texttt{GPT-4o} \cite{OpenAI2023GPT4}, \texttt{LLaVA-v1.6-mistral-7b} \cite{liu2023visual},
\texttt{Qwen2.5-VL-7B-Instruct} \cite{bai2025qwen2}, and \texttt{Molmo-7B-D-0924} \cite{deitke2024molmo}. We will use \texttt{GPT-4o}, \texttt{LLaVA-v1.6}, \texttt{Qwen2.5} and \texttt{Molmo-7B} as abbreviation. Additionally, we propose a framework for intent prediction, SGG-Intent to serve as the baseline on this benchmark for future works.

\subsection{Evaluation Tasks and Metrics}

\begin{itemize}[leftmargin=8pt, itemsep=1pt, parsep=1pt, topsep=0pt, partopsep=0pt]
    \item \textbf{Object Detection (OD)} For $m$ predicted objects and $n$ objects in the ground-truth sample, the detection accuracy corresponds to only the $n$ objects and corresponding matches where IoU greater than set threshold.

    \item \textbf{Intent Prediction (IP)} We decompose this task into two components: Lateral Intent Prediction (LIP), Vertical Intent Prediction (VIP). We report the individual and combined accuracies for these components.

    \item \textbf{Risk Assessment (RA)} The risk score annotations are binary and heavily imbalanced towards positive risk. We therefore report a balanced accuracy (BA) for this task along with the F1 score.

    \item \textbf{Action Suggestion (AS)} Recommendations are free‐form text, so we adopt BERTScore to evaluate this task. We report the F1 score obtained from the BERTScore of each sample.

\end{itemize}

\subsection{Evaluation}

This section presents evaluation results on the DRAMA-X benchmark, focusing on task level and model level results. The results are aggregated in Table~\ref{eval-results}.
\begin{table}
\small
\setlength{\tabcolsep}{9pt}
  \caption{Evaluation results of various VLMs and SGG-Intent frameworks across benchmark tasks: Object Detection (OD), Intent Prediction (IP), Risk Assessment (RA), and Action Suggestion (AS). All results are represented by accuracy in percentages (\%)}
  \label{eval-results}
  \centering
  \begin{tabular}{l c c c c c c c}
    \toprule
    \textbf{Method}
      & \textbf{OD}$\uparrow$
      & \multicolumn{3}{c}{\textbf{Intent Prediction (IP)}$\uparrow$}
      & \multicolumn{2}{c}{\textbf{RA}$\uparrow$}
      & \textbf{AS}$\uparrow$ \\
    \cmidrule(lr){3-5} \cmidrule(lr){6-7}
      & 
      & \textbf{LIP} 
      & \textbf{VIP} 
      & \textbf{Combined} 
      & \textbf{BA} 
      & \textbf{F1} 
      &  \\
    \midrule
    \multicolumn{8}{c}{\textbf{w/o}\ \texttt{SSG-Intent}}\\
    \midrule
    \texttt{Molmo-7B}               & 0.00 & 0.00 & 0.00 & 0.00 & 52.09 & 69.63  & 82.09 \\
    \texttt{LLaVA-v1.6}        & 0.05 & 0.00 & 0.00 & 0.00& 46.13
  & 66.75 & 80.14 \\
    \texttt{GPT-4o}                       & 4.54 & 0.08 & 0.09 & 0.09 & 57.00 & 28.54 & 82.64 \\
    \texttt{Gemini2.5-Pro}        & 0.00 & 0.00 & 0.00 & 0.00 & 49.44 & 67.82 & 80.20 \\
    \texttt{Qwen2.5}       & 38.04 & 15.72 & 11.22 & 13.47 & 50.52 & 66.70 & 82.56 \\
    \midrule
    \multicolumn{8}{c}{+\ \texttt{SSG-Intent}}\\
    \midrule
    \texttt{Molmo-7B} & 0.10 & 0.00 & 0.00 & 0.00 & 51.79 & 93.96 & 82.51 \\
    \texttt{LLaVA-v1.6}  & 0.34 & 0.00 & 0.00 & 0.00 & \textbf{55.32} & 64.75 & 82.43 \\
                                          
    \texttt{GPT-4o}        & 2.18 & 0.04 & 0.05 & 0.05 & 52.50 & 91.64 & 82.40 \\
    \texttt{Gemini2.5-Pro}        & 1.02 & 0.02 & 0.01 & 0.02 & 52.22 & 94.45 & 82.52 \\
    \texttt{Qwen2.5}
                                          & \textbf{48.53} & \textbf{18.51} & \textbf{15.84} & \textbf{17.18} & 48.79 & \textbf{95.41} & \textbf{84.21} \\
    \bottomrule
  \end{tabular}
\end{table}


\paragraph{Direct Evaluation of VLMs.}
When evaluating VLMs directly on our benchmark tasks without any structural framework, we observe consistently poor performance across all models. This direct evaluation serves as a baseline to understand the inherent limitations of current VLMs for complex scene understanding tasks.
Looking at the top-half of Table~\ref{eval-results}, standalone VLMs achieve significantly lower scores across all tasks compared to their framework-enhanced counterparts. Notably \texttt{Molmo-7B} and \texttt{LLava-v1.6} are unable to detect and determine the intent labels for any object. There is no effect of the performance of OD and IP tasks on RA task.

\paragraph{Evaluating Object Localization Capability of VLMs}
Our experiments reveal a critical limitation of current VLMs: while they can often recognize objects present in a scene, they struggle significantly with precise object localization. Among the evaluated models, \texttt{Qwen2.5} demonstrates notably superior performance in object detection (48.53\%), outperforming other models including GPT-4o (2.18\%), LLaVA (0.34\%), and Molmo (0.1\%).
The poor object localization capability stems from VLMs' training objectives, which typically prioritize general scene understanding and question-answering capabilities over precise spatial reasoning. When prompted to provide bounding box coordinates, most models produce imprecise boundaries that fail to accurately capture object locations, significantly impacting downstream tasks.

\begin{wraptable}{r}{0.5\textwidth}
\small
\setlength{\tabcolsep}{7pt}
  \caption{Intent Prediction evaluation results with added object detection ground truth prior.}
  \label{eval2-results}
  \centering
  \resizebox{\linewidth}{!}{
  \begin{tabular}{lccc}
    \toprule
    \textbf{Method}  
      & \multicolumn{3}{c}{\textbf{Intent Prediction (IP)} $\uparrow$} 
      \\
    \cmidrule(lr){2-4}
     & \textbf{LIP} & \textbf{VIP} & \textbf{Combined}\\
    \midrule
    
    \texttt{Molmo}  & 1.82 & 1.14 & 1.48  \\
    \texttt{LLaVA-v1.6}  & 23.93 & 37.61 & 10.25  \\
    \texttt{GPT-4o}  & 19.03 & 23.63 & 21.33  \\
    \texttt{Gemini2.5-Pro}         & 11.32 & 9.85 & 10.59  \\
    \texttt{Qwen2.5}  & \textbf{45.11} & \textbf{38.36} & \textbf{41.73} \\
    \bottomrule
  \end{tabular}}
\end{wraptable}

\paragraph{Intent Prediction.}
Intent prediction performance demonstrates a strong correlation with object detection accuracy. While the VLMs integrated with our SGG-Intent framework (shown in the bottom half of Table~\ref{eval-results}) exhibit notable improvements over standalone VLM performance, their effectiveness remains constrained by imprecise object localization. This effect stands out in the case of GPT-4o with the combined intent prediction accuracy < 1\%. To isolate this dependency and establish an upper bound on intent prediction capability, we conducted additional experiments providing ground truth bounding boxes as prior information to the intent generation process.
The results presented in Table~\ref{eval2-results} reveal substantial performance gains, with \texttt{Qwen2.5} achieving a remarkable 202\% relative improvement in combined intent accuracy when leveraging ground truth object locations. This dramatic enhancement quantitatively confirms our hypothesis that accurate object detection serves as a critical foundation for downstream intent prediction. Nevertheless, even with perfect object localization, the overall intent prediction performance remains suboptimal, particularly in the context of safety-critical scenarios involving vulnerable road users. These findings underscore the significant gap in current VLMs' ability to interpret nuanced human movement intentions and highlight the urgent need for specialized methods focused on pedestrian and cyclist intent understanding in autonomous driving contexts.

\paragraph{Risk Assessment.}
The risk assessment phase builds upon preceding analytical components, necessitating both precise object detection and accurate intent prediction capabilities. Quantitative evaluation reveals that standalone Vision-Language Models (VLMs) demonstrate constrained performance in risk assessment tasks with a distinct jump in F1 scores for SGG-Intent enhanced task results.
It is noteworthy that errors from object-level tasks do not directly propagate to the scene-level risk assessment framework. This empirical observation suggests that a sufficiently comprehensive scene graph representation can yield adequate risk assessment metrics, thereby emphasizing the critical importance of robust scene graph generation as a foundational capability. Our analysis indicates a minor improvement with \texttt{Qwen2.5} compared to baseline models when evaluated using the F1 metric. However, we simultaneously observe a reduction in Balanced Accuracy (BA), indicating a decreased propensity to generate negative risk annotations. Given that the accurate identification of potentially hazardous scenarios represents the primary objective of the system, the performance characteristics demonstrated by Qwen remain superior within the evaluated context despite this trade-off.

\paragraph{Suggested Action.}
For the action suggestion task, which represents the culmination of our sequential pipeline, we observe once again that accurate scene representation is crucial. We observe comparable results across all models and methods.
With the SGG-Intent framework providing structured information from previous tasks, action suggestion quality improves marginally across all models. Qwen with SGG-Intent achieves the highest BERTScore of 84.21\%, indicating that its generated action suggestions align most closely with ground truth suggestions from human annotators.

\subsection{Ablation Studies}
To understand the sequential dependency between tasks in our framework, we conducted ablation studies focusing on the effect of task execution order and the impact of providing oracle information at different stages. We choose \texttt{Qwen2.5} as the base model for this study. We tested the results on three different setups: 1) Ground Truth (GT) input at every stage, 2) Only sequential model outputs enhanced with SGG-Intent, 3) Direct prediction with no prior.

\begin{table}[h]
  \small
  \caption{Ablation study results assessing the impact of scene-graph priors on benchmark tasks.}
  \label{tab:ablation}
  \centering
  \begin{tabular}{l c c c c c c c}
    \toprule
    \textbf{Method}
      & \textbf{OD}$\uparrow$
      & \multicolumn{3}{c}{\textbf{Intent Prediction (IP)}$\uparrow$}
      & \multicolumn{2}{c}{\textbf{RA}$\uparrow$}
      & \textbf{AS }$\uparrow$ \\
    \cmidrule(lr){3-5} \cmidrule(lr){6-7}
      & 
      & \textbf{LIP} 
      & \textbf{VIP} 
      & \textbf{Combined} 
      & \textbf{BA} 
      & \textbf{F1} 
      &  \\
      \midrule
    w/ GT                        & - & \textbf{45.11} & \textbf{38.36} & \textbf{41.73} & \textbf{54.60} & \textbf{96.65} & 82.36\\
    w/ model preds       & 48.53 & 18.51 & 15.84 & 17.18 & 48.79 & 95.41 & \textbf{84.21} \\
    No prior                       & 38.04 & 15.72 & 11.22 & 13.47 & 50.52 & 66.70 & 82.56\\
    \bottomrule
  \end{tabular}
\end{table}

\paragraph{Effect of the Sequence of Task Execution.} We investigated how performance changes when bypassing certain steps in our pipeline or altering the sequence of execution specifically for \texttt{Qwen2.5}. Our experiments confirm that the proposed sequential approach (OD → IP → RA → AS) is optimal. When models skip the object detection step and attempt intent prediction directly, performance drops by 21.6\%. Ground truth intent labels enhance risk assessment F1 score by  1.14\% over risk scores generated with inaccurate intent priors. The ablation studies further validate our framework design and highlight the importance of structured, sequential reasoning for complex visual understanding tasks involving human intent inference and risk assessment.


\section{Conclusion}
\label{sec:conclusion}
This paper introduces DRAMA-X, a novel benchmark for fine-grained intent prediction of vulnerable road users in high-risk driving scenarios. By focusing on pedestrians and cyclists in safety-critical situations, we address a critical gap in autonomous driving research. Our SGG-Intent framework demonstrates that structured, compositional reasoning significantly outperforms direct VLM querying across all benchmark tasks. Experimental results reveal a critical bottleneck: current VLMs struggle with precise object localization, creating cascading errors that amplify through subsequent reasoning stages.
\paragraph{Limitations and Future Work} Despite promising results, our automated annotation pipeline may introduce errors in complex scenes with multiple VRUs. The benchmark's single-frame approach limits temporal understanding, and we rely on general VLMs rather than specialized scene graph models. DRAMA primarily addresses high-risk scenarios but doesn't cover the full spectrum of abnormal risks.
Future work should include diverse risk levels and scenario categorization by VRU involvement. Incorporating specialized models and temporal reasoning would improve performance in ambiguous situations. Human-in-the-loop refinement could enhance ground truth quality. Other directions include generating explanations for predicted intents and exploring multi-modal fusion for adverse conditions. While contributing to road safety, we acknowledge the importance of testing across diverse populations and implementing human oversight for real-world deployment.
DRAMA-X advances autonomous driving safety by highlighting VRU intent prediction challenges and establishing a rigorous evaluation framework. Our findings demonstrate the need for specialized approaches where general-purpose VLMs currently underperform.





\medskip

{\small
\bibliographystyle{unsrtnat}
\bibliography{references}

\begin{thebibliography}{66}
\providecommand{\natexlab}[1]{#1}
\providecommand{\url}[1]{\texttt{#1}}
\expandafter\ifx\csname urlstyle\endcsname\relax
  \providecommand{\doi}[1]{doi: #1}\else
  \providecommand{\doi}{doi: \begingroup \urlstyle{rm}\Url}\fi

\bibitem[Najibi et~al.(2023)Najibi, Ji, Zhou, Qi, Yan, Ettinger, and Anguelov]{najibi2023unsupervised}
Mahyar Najibi, Jingwei Ji, Yin Zhou, Charles~R Qi, Xinchen Yan, Scott Ettinger, and Dragomir Anguelov.
\newblock Unsupervised 3d perception with 2d vision-language distillation for autonomous driving.
\newblock In \emph{Proceedings of the IEEE/CVF International Conference on Computer Vision}, pages 8602--8612, 2023.

\bibitem[Yao et~al.(2023)Yao, Han, Liang, Xu, Zhang, Li, and Xu]{yao2023detclipv2}
Lewei Yao, Jianhua Han, Xiaodan Liang, Dan Xu, Wei Zhang, Zhenguo Li, and Hang Xu.
\newblock Detclipv2: Scalable open-vocabulary object detection pre-training via word-region alignment.
\newblock In \emph{Proceedings of the IEEE/CVF Conference on Computer Vision and Pattern Recognition}, pages 23497--23506, 2023.

\bibitem[Tian et~al.(2024)Tian, Gu, Li, Liu, Wang, Zhao, Zhan, Jia, Lang, and Zhao]{tian2024drivevlm}
Xiaoyu Tian, Junru Gu, Bailin Li, Yicheng Liu, Yang Wang, Zhiyong Zhao, Kun Zhan, Peng Jia, Xianpeng Lang, and Hang Zhao.
\newblock Drivevlm: The convergence of autonomous driving and large vision-language models.
\newblock \emph{arXiv preprint arXiv:2402.12289}, 2024.

\bibitem[Gao et~al.(2025{\natexlab{a}})Gao, Wu, Wang, Liu, Zhou, and Tu]{gao2025langcoop}
Xiangbo Gao, Yuheng Wu, Rujia Wang, Chenxi Liu, Yang Zhou, and Zhengzhong Tu.
\newblock Langcoop: Collaborative driving with language.
\newblock In \emph{Proceedings of the Computer Vision and Pattern Recognition Conference}, pages 4226--4237, 2025{\natexlab{a}}.

\bibitem[Wang et~al.(2025)Wang, Xing, Can, Li, Hua, Tian, Mo, Gao, Wu, Zhou, et~al.]{wang2025generative}
Yuping Wang, Shuo Xing, Cui Can, Renjie Li, Hongyuan Hua, Kexin Tian, Zhaobin Mo, Xiangbo Gao, Keshu Wu, Sulong Zhou, et~al.
\newblock Generative ai for autonomous driving: Frontiers and opportunities.
\newblock \emph{arXiv preprint arXiv:2505.08854}, 2025.

\bibitem[Xing et~al.(2024)Xing, Hua, Gao, Zhu, Li, Tian, Li, Huang, Yang, Wang, et~al.]{xing2024autotrust}
Shuo Xing, Hongyuan Hua, Xiangbo Gao, Shenzhe Zhu, Renjie Li, Kexin Tian, Xiaopeng Li, Heng Huang, Tianbao Yang, Zhangyang Wang, et~al.
\newblock Autotrust: Benchmarking trustworthiness in large vision language models for autonomous driving.
\newblock \emph{arXiv preprint arXiv:2412.15206}, 2024.

\bibitem[Luo et~al.(2025)Luo, Yang, Ding, Gao, Xing, Zhou, Tu, and Liu]{luo2025v2x}
Xuewen Luo, Fengze Yang, Fan Ding, Xiangbo Gao, Shuo Xing, Yang Zhou, Zhengzhong Tu, and Chenxi Liu.
\newblock V2x-unipool: Unifying multimodal perception and knowledge reasoning for autonomous driving.
\newblock \emph{arXiv preprint arXiv:2506.02580}, 2025.

\bibitem[Gao et~al.(2025{\natexlab{b}})Gao, Wu, Zhang, Tian, Zhou, and Tu]{gao2025automated}
Xiangbo Gao, Keshu Wu, Hao Zhang, Kexin Tian, Yang Zhou, and Zhengzhong Tu.
\newblock Automated vehicles should be connected with natural language.
\newblock \emph{arXiv preprint arXiv:2507.01059}, 2025{\natexlab{b}}.

\bibitem[Zheng et~al.(2024)Zheng, Zhao, Gong, Zhu, and Wu]{zheng2024simplellm4ad}
Peiru Zheng, Yun Zhao, Zhan Gong, Hong Zhu, and Shaohua Wu.
\newblock Simplellm4ad: An end-to-end vision-language model with graph visual question answering for autonomous driving.
\newblock \emph{arXiv preprint arXiv:2407.21293}, 2024.

\bibitem[Xu et~al.(2024)Xu, Zhang, Xie, Zhao, Guo, Wong, Li, and Zhao]{xu2024drivegpt4}
Zhenhua Xu, Yujia Zhang, Enze Xie, Zhen Zhao, Yong Guo, Kwan-Yee~K Wong, Zhenguo Li, and Hengshuang Zhao.
\newblock Drivegpt4: Interpretable end-to-end autonomous driving via large language model.
\newblock \emph{IEEE Robotics and Automation Letters}, 2024.

\bibitem[You et~al.(2024)You, Shi, Jiang, Huang, Gan, Wu, Cheng, Li, and Ran]{you2024v2x}
Junwei You, Haotian Shi, Zhuoyu Jiang, Zilin Huang, Rui Gan, Keshu Wu, Xi~Cheng, Xiaopeng Li, and Bin Ran.
\newblock V2x-vlm: End-to-end v2x cooperative autonomous driving through large vision-language models.
\newblock \emph{arXiv preprint arXiv:2408.09251}, 2024.

\bibitem[Jiang et~al.(2024)Jiang, Chen, Liao, Zhang, Yin, Zhang, Huang, Liu, and Wang]{jiang2024senna}
Bo~Jiang, Shaoyu Chen, Bencheng Liao, Xingyu Zhang, Wei Yin, Qian Zhang, Chang Huang, Wenyu Liu, and Xinggang Wang.
\newblock Senna: Bridging large vision-language models and end-to-end autonomous driving.
\newblock \emph{arXiv preprint arXiv:2410.22313}, 2024.

\bibitem[Mao et~al.(2024)Mao, Qian, Ye, Zhao, and Wang]{Mao2023GPTDriver}
Jiageng Mao, Yuxi Qian, Junjie Ye, Hang Zhao, and Yue Wang.
\newblock Gpt-driver: Learning to drive with gpt.
\newblock In \emph{International Conference on Learning Representations (ICLR)}, 2024.
\newblock arXiv:2310.01415.

\bibitem[Abu~Tami et~al.(2024)Abu~Tami, Ashqar, and Elhenawy]{Tami2024MLLMEvents}
Mohammad Abu~Tami, Huthaifa~I. Ashqar, and Mohammed Elhenawy.
\newblock Using multimodal large language models for automated detection of traffic safety-critical events.
\newblock \emph{arXiv preprint arXiv:2406.13894}, 2024.

\bibitem[Xing et~al.(2025)Xing, Qian, Wang, Hua, Tian, Zhou, and Tu]{xing2025openemma}
Shuo Xing, Chengyuan Qian, Yuping Wang, Hongyuan Hua, Kexin Tian, Yang Zhou, and Zhengzhong Tu.
\newblock Openemma: Open-source multimodal model for end-to-end autonomous driving.
\newblock In \emph{Proceedings of the Winter Conference on Applications of Computer Vision}, pages 1001--1009, 2025.

\bibitem[Kutela et~al.(2022)Kutela, Das, and Dadashova]{kutela2022mining}
Boniphace Kutela, Subasish Das, and Bahar Dadashova.
\newblock Mining patterns of autonomous vehicle crashes involving vulnerable road users to understand the associated factors.
\newblock \emph{Accident Analysis \& Prevention}, 165:\penalty0 106473, 2022.

\bibitem[Abdel-Aty and Ding(2024)]{abdel2024matched}
Mohamed Abdel-Aty and Shengxuan Ding.
\newblock A matched case-control analysis of autonomous vs human-driven vehicle accidents.
\newblock \emph{Nature communications}, 15\penalty0 (1):\penalty0 4931, 2024.

\bibitem[Li et~al.(2025)Li, Cao, Gao, Tian, Wu, Anis, Zhang, Long, Jiang, Li, et~al.]{li2025simulating}
Zihao Li, Xinyuan Cao, Xiangbo Gao, Kexin Tian, Keshu Wu, Mohammad Anis, Hao Zhang, Keke Long, Jiwan Jiang, Xiaopeng Li, et~al.
\newblock Simulating the unseen: Crash prediction must learn from what did not happen.
\newblock \emph{arXiv preprint arXiv:2505.21743}, 2025.

\bibitem[Rasouli et~al.(2017)Rasouli, Kotseruba, and Tsotsos]{Rasouli2017JAAD}
Amir Rasouli, Iuliia Kotseruba, and John~K. Tsotsos.
\newblock Are they going to cross? a benchmark dataset and baseline for pedestrian crosswalk behavior.
\newblock In \emph{Proc. IEEE Int. Conf. on Computer Vision Workshops (ICCVW)}, 2017.

\bibitem[Rasouli et~al.(2019)Rasouli, Kotseruba, Kunic, and Tsotsos]{Rasouli2019PIE}
Amir Rasouli, Iuliia Kotseruba, Toni Kunic, and John~K. Tsotsos.
\newblock {PIE}: A large-scale dataset and models for pedestrian intention estimation and trajectory prediction.
\newblock In \emph{Proc. IEEE Int. Conf. on Computer Vision (ICCV)}, 2019.

\bibitem[Sadeghian et~al.(2019)Sadeghian, Kosaraju, Sadeghian, Hirose, Rezatofighi, and Savarese]{sadeghian2019sophie}
Amir Sadeghian, Vineet Kosaraju, Ali Sadeghian, Noriaki Hirose, Hamid Rezatofighi, and Silvio Savarese.
\newblock Sophie: An attentive gan for predicting paths compliant to social and physical constraints.
\newblock In \emph{Proceedings of the IEEE/CVF conference on computer vision and pattern recognition}, pages 1349--1358, 2019.

\bibitem[Alahi et~al.(2016)Alahi, Goel, Ramanathan, Robicquet, Fei-Fei, and Savarese]{alahi2016social}
Alexandre Alahi, Kratarth Goel, Vignesh Ramanathan, Alexandre Robicquet, Li~Fei-Fei, and Silvio Savarese.
\newblock Social lstm: Human trajectory prediction in crowded spaces.
\newblock In \emph{Proceedings of the IEEE conference on computer vision and pattern recognition}, pages 961--971, 2016.

\bibitem[Zhou et~al.(2021)Zhou, Wu, Cheng, Qi, Hu, Kang, and Zheng]{zhou2021social}
Yutao Zhou, Huayi Wu, Hongquan Cheng, Kunlun Qi, Kai Hu, Chaogui Kang, and Jie Zheng.
\newblock Social graph convolutional lstm for pedestrian trajectory prediction.
\newblock \emph{IET Intelligent Transport Systems}, 15\penalty0 (3):\penalty0 396--405, 2021.

\bibitem[Ling et~al.(2024)Ling, Ma, Zhang, Xie, and Weng]{ling2024pedast}
Yancheng Ling, Zhenliang Ma, Qi~Zhang, Bangquan Xie, and Xiaoxiong Weng.
\newblock Pedast-gcn: Fast pedestrian crossing intention prediction using spatial--temporal attention graph convolution networks.
\newblock \emph{IEEE Transactions on Intelligent Transportation Systems}, 2024.

\bibitem[Malla et~al.(2023)Malla, Choi, Dwivedi, Choi, and Li]{malla2023drama}
Srikanth Malla, Chiho Choi, Isht Dwivedi, Joon~Hee Choi, and Jiachen Li.
\newblock Drama: Joint risk localization and captioning in driving.
\newblock In \emph{Proceedings of the IEEE/CVF winter conference on applications of computer vision}, pages 1043--1052, 2023.

\bibitem[Bai et~al.(2025)Bai, Chen, Liu, Wang, Ge, Song, Dang, Wang, Wang, Tang, et~al.]{bai2025qwen2}
Shuai Bai, Keqin Chen, Xuejing Liu, Jialin Wang, Wenbin Ge, Sibo Song, Kai Dang, Peng Wang, Shijie Wang, Jun Tang, et~al.
\newblock Qwen2. 5-vl technical report.
\newblock \emph{arXiv preprint arXiv:2502.13923}, 2025.

\bibitem[{OpenAI}(2023)]{OpenAI2023GPT4}
{OpenAI}.
\newblock Gpt-4 technical report.
\newblock \emph{arXiv:2303.08774}, 2023.

\bibitem[Deitke et~al.(2024)Deitke, Clark, Lee, Tripathi, Yang, Park, Salehi, Muennighoff, Lo, Soldaini, et~al.]{deitke2024molmo}
Matt Deitke, Christopher Clark, Sangho Lee, Rohun Tripathi, Yue Yang, Jae~Sung Park, Mohammadreza Salehi, Niklas Muennighoff, Kyle Lo, Luca Soldaini, et~al.
\newblock Molmo and pixmo: Open weights and open data for state-of-the-art multimodal models.
\newblock \emph{arXiv preprint arXiv:2409.17146}, 2024.

\bibitem[Li et~al.(2024{\natexlab{a}})Li, Zhang, Guo, Zhang, Li, Zhang, Zhang, Zhang, Li, Liu, et~al.]{li2024llava}
Bo~Li, Yuanhan Zhang, Dong Guo, Renrui Zhang, Feng Li, Hao Zhang, Kaichen Zhang, Peiyuan Zhang, Yanwei Li, Ziwei Liu, et~al.
\newblock Llava-onevision: Easy visual task transfer.
\newblock \emph{arXiv preprint arXiv:2408.03326}, 2024{\natexlab{a}}.

\bibitem[Kooij et~al.(2014)Kooij, Schneider, Flohr, and Gavrila]{Kooij2014Context}
Julien F.~P. Kooij, N.~Schneider, F.~Flohr, and Dariu~M. Gavrila.
\newblock Context-based pedestrian path prediction.
\newblock \emph{IEEE Trans. Pattern Analysis and Machine Intelligence (PAMI)}, 36\penalty0 (6):\penalty0 1242--1257, 2014.

\bibitem[Schneemann and Heinemann(2016)]{Schneemann2016Context}
Friederike Schneemann and Patrick Heinemann.
\newblock Context-based detection of pedestrian crossing intention for autonomous driving in urban environments.
\newblock In \emph{Proc. IEEE Int. Conf. on Intelligent Transportation Systems (ITSC)}, pages 2793--2798, 2016.

\bibitem[Fang and Lopez(2018)]{Fang2018}
Zhijie Fang and Antonio~M. Lopez.
\newblock Is the pedestrian going to cross? answering by 2d pose estimation.
\newblock In \emph{IEEE Intelligent Vehicles Symposium (IV)}, 2018.

\bibitem[Gao et~al.(2024)Gao, Kanu-Asiegbu, and Du]{gao2024mambast}
Xiangbo Gao, Asiegbu~Miracle Kanu-Asiegbu, and Xiaoxiao Du.
\newblock Mambast: A plug-and-play cross-spectral spatial-temporal fuser for efficient pedestrian detection.
\newblock In \emph{2024 IEEE 27th International Conference on Intelligent Transportation Systems (ITSC)}, pages 2027--2034. IEEE, 2024.

\bibitem[Kotseruba et~al.(2021)Kotseruba, Rasouli, and Tsotsos]{Kotseruba2021Review}
Iuliia Kotseruba, Amir Rasouli, and John~K. Tsotsos.
\newblock Pedestrian intention prediction in autonomous driving: A review of data, methods, and evaluations.
\newblock \emph{arXiv:2105.04149}, 2021.

\bibitem[Gao et~al.(2025{\natexlab{c}})Gao, Wu, Luo, Wu, Chen, Wang, Liu, Zhou, and Tu]{gao2025airv2x}
Xiangbo Gao, Yuheng Wu, Xuewen Luo, Keshu Wu, Xinghao Chen, Yuping Wang, Chenxi Liu, Yang Zhou, and Zhengzhong Tu.
\newblock Airv2x: Unified air-ground vehicle-to-everything collaboration.
\newblock \emph{arXiv preprint arXiv:2506.19283}, 2025{\natexlab{c}}.

\bibitem[Geiger et~al.(2013)Geiger, Lenz, Stiller, and Urtasun]{Geiger2013KITTI}
Andreas Geiger, Philip Lenz, Christoph Stiller, and Raquel Urtasun.
\newblock Vision meets robotics: The {KITTI} dataset.
\newblock \emph{Int. Journal of Robotics Research}, 32\penalty0 (11):\penalty0 1231--1237, 2013.

\bibitem[Caesar et~al.(2020)Caesar, Bankiti, Lang, Vora, Liong, Xu, Krishnan, Pan, Baldan, and Beijbom]{Caesar2020nuScenes}
Holger Caesar, Varun Bankiti, Alex~H. Lang, Sourabh Vora, Venice~Erin Liong, Qiang Xu, Anush Krishnan, Yu~Pan, Giancarlo Baldan, and Oscar Beijbom.
\newblock {nuScenes}: A multimodal dataset for autonomous driving.
\newblock In \emph{Proc. IEEE Conf. on Computer Vision and Pattern Recognition (CVPR)}, 2020.

\bibitem[Sun et~al.(2020)Sun, Kretzschmar, Dotiwalla, Chouard, Patnaik, and etc.]{Sun2020Waymo}
Pei Sun, Henrik Kretzschmar, Xerxes Dotiwalla, Aurelien Chouard, Vijaysai Patnaik, and etc.
\newblock Scalability in perception for autonomous driving: Waymo open dataset.
\newblock In \emph{Proc. IEEE/CVF Conf. on Computer Vision and Pattern Recognition (CVPR)}, 2020.

\bibitem[Chang et~al.(2019)Chang, Lambert, Sangkloy, Singh, Bak, Hartnett, Ramanan, and Hays]{Chang2019Argoverse}
Ming-Fang Chang, John Lambert, Patsorn Sangkloy, Jagjeet Singh, Slawomir Bak, Andrew Hartnett, Deva Ramanan, and James Hays.
\newblock Argoverse: 3d tracking and forecasting with rich maps.
\newblock In \emph{Proc. IEEE/CVF Conf. on Computer Vision and Pattern Recognition (CVPR)}, 2019.

\bibitem[Yu et~al.(2020)Yu, Chen, Wang, Xian, Chen, Liu, Madhavan, and Darrell]{Yu2020BDD100K}
Fisher Yu, Haofeng Chen, Xin Wang, Wenqi Xian, Yingying Chen, Fangchen Liu, Vashisht Madhavan, and Trevor Darrell.
\newblock {BDD100K}: A diverse driving dataset for heterogeneous multitask learning.
\newblock In \emph{Proc. IEEE/CVF Conf. on Computer Vision and Pattern Recognition (CVPR)}, 2020.

\bibitem[Ettinger et~al.(2021)Ettinger, Cheng, Caine, Liu, Zhao, Pradhan, Chai, Sapp, Qi, Zhou, et~al.]{ettinger2021large}
Scott Ettinger, Shuyang Cheng, Benjamin Caine, Chenxi Liu, Hang Zhao, Sabeek Pradhan, Yuning Chai, Ben Sapp, Charles~R Qi, Yin Zhou, et~al.
\newblock Large scale interactive motion forecasting for autonomous driving: The waymo open motion dataset.
\newblock In \emph{Proceedings of the IEEE/CVF International Conference on Computer Vision}, pages 9710--9719, 2021.

\bibitem[Sima et~al.(2024)Sima, Renz, Chitta, Chen, Zhang, Xie, Beiswenger, Luo, Geiger, and Li]{Sima2024DriveLM}
Chonghao Sima, Katrin Renz, Kashyap Chitta, Li~Chen, Hanxue Zhang, Chengen Xie, Jens Beiswenger, Ping Luo, Andreas Geiger, and Hongyang Li.
\newblock {DriveLM}: Driving with graph visual question answering.
\newblock In \emph{Proc. European Conf. on Computer Vision (ECCV)}, 2024.

\bibitem[Li et~al.(2024{\natexlab{b}})Li, Ge, Li, Xu, Tomizuka, Tang, Ding, and Zhan]{Li2024WOMD-Reasoning}
Yiheng Li, Chongjian Ge, Chenran Li, Chenfeng Xu, Masayoshi Tomizuka, Chen Tang, Mingyu Ding, and Wei Zhan.
\newblock {WOMD-Reasoning}: A large-scale language dataset for interaction and driving intentions reasoning.
\newblock \emph{arXiv:2407.04281}, 2024{\natexlab{b}}.

\bibitem[Kim et~al.(2018)Kim, Misu, Chen, Tawari, and Canny]{Kim2018BDD-X}
Jinkyu Kim, Teruhisa Misu, Yi-Ting Chen, Ashish Tawari, and John Canny.
\newblock Textual explanations for self-driving vehicles.
\newblock In \emph{Proc. European Conf. on Computer Vision (ECCV)}, 2018.

\bibitem[Xu and et~al.(2020)]{Xu2020OIA}
Yiran Xu and et~al.
\newblock Explainable object-induced action decision for autonomous vehicles.
\newblock In \emph{Proc. IEEE/CVF Conf. on Computer Vision and Pattern Recognition (CVPR)}, 2020.

\bibitem[Nie et~al.(2024)Nie, Peng, Wang, Cai, Han, Xu, and Zhang]{Nie2024Reason2Drive}
Ming Nie, Renyuan Peng, Chunwei Wang, Xinyue Cai, Jianhua Han, Hang Xu, and Li~Zhang.
\newblock {Reason2Drive}: Towards interpretable and chain-based reasoning for autonomous driving.
\newblock In \emph{Proc. European Conf. on Computer Vision (ECCV)}, 2024.

\bibitem[Krishna et~al.(2017)Krishna, Zhu, Groth, Johnson, Hata, Kravitz, Chen, Kalantidis, Li, Shamma, Bernstein, and Fei-Fei]{Krishna2017VG}
Ranjay Krishna, Yuke Zhu, Oliver Groth, Justin Johnson, Kenji Hata, Jack Kravitz, Stephanie Chen, Yannis Kalantidis, Li-Jia Li, David~A. Shamma, Michael Bernstein, and Li~Fei-Fei.
\newblock Visual genome: Connecting language and vision using crowdsourced dense image annotations.
\newblock \emph{International Journal of Computer Vision}, 123\penalty0 (1):\penalty0 32--73, 2017.

\bibitem[Xu et~al.(2017)Xu, Zhu, Choy, and Fei-Fei]{Xu2017SceneGraph}
Danfei Xu, Yuke Zhu, Christopher~B. Choy, and Li~Fei-Fei.
\newblock Scene graph generation by iterative message passing.
\newblock In \emph{Proceedings of the IEEE Conference on Computer Vision and Pattern Recognition (CVPR)}, 2017.

\bibitem[Zellers et~al.(2018)Zellers, Yatskar, Thomson, and Choi]{Zellers2018NeuralMotifs}
Rowan Zellers, Mark Yatskar, Sam Thomson, and Yejin Choi.
\newblock Neural motif networks for visual scene graph generation.
\newblock In \emph{Proceedings of the IEEE Conference on Computer Vision and Pattern Recognition (CVPR)}, 2018.

\bibitem[Mitra et~al.(2023)Mitra, Huang, Darrell, and Herzig]{Mitra2023CCoT}
Chancharik Mitra, Brandon Huang, Trevor Darrell, and Roei Herzig.
\newblock Compositional chain-of-thought prompting for large multimodal models.
\newblock \emph{arXiv preprint arXiv:2311.17076}, 2023.

\bibitem[Elskhawy et~al.(2025)Elskhawy, Li, Navab, and Busam]{Elskhawy2025PRISM}
Abdelrahman Elskhawy, Mengze Li, Nassir Navab, and Benjamin Busam.
\newblock Prism-0: A predicate-rich scene graph generation framework for zero-shot open-vocabulary tasks.
\newblock \emph{arXiv preprint arXiv:2504.00844}, 2025.

\bibitem[Yin et~al.(2024)Yin, Xu, Wu, Zhou, and Lu]{Yin2024SGNav}
Hang Yin, Xiuwei Xu, Zhenyu Wu, Jie Zhou, and Jiwen Lu.
\newblock Sg-nav: Online 3d scene graph prompting for llm-based zero-shot object navigation.
\newblock In \emph{NeurIPS 2024}, 2024.

\bibitem[Li et~al.(2024{\natexlab{c}})Li, Zhang, Lin, Chen, and He]{Li2024PixelsToGraphs}
Rongjie Li, Songyang Zhang, Dahua Lin, Kai Chen, and Xuming He.
\newblock From pixels to graphs: Open-vocabulary scene graph generation with vision-language models.
\newblock In \emph{Proceedings of the IEEE/CVF Conference on Computer Vision and Pattern Recognition (CVPR)}, 2024{\natexlab{c}}.

\bibitem[Chen et~al.(2024)Chen, Wu, Lei, Zhang, and Chen]{Chen2024OvSGTR}
Zuyao Chen, Jinlin Wu, Zhen Lei, Zhaoxiang Zhang, and Changwen Chen.
\newblock Expanding scene graph boundaries: Fully open-vocabulary scene graph generation via visual-concept alignment and retention (ovsgtr).
\newblock In \emph{Proceedings of the European Conference on Computer Vision (ECCV)}, 2024.

\bibitem[Im et~al.(2024)Im, Nam, Park, Lee, and Park]{Im2024EGTR}
Jinbae Im, JeongYeon Nam, Nokyung Park, Hyungmin Lee, and Seunghyun Park.
\newblock Egtr: Extracting graph from transformer for scene graph generation.
\newblock In \emph{Proceedings of the IEEE/CVF Conference on Computer Vision and Pattern Recognition (CVPR)}, 2024.

\bibitem[Liu et~al.(2025)Liu, Li, Wang, and He]{Liu2024RAHP}
Tao Liu, Rongjie Li, Chongyu Wang, and Xuming He.
\newblock Relation-aware hierarchical prompt for open-vocabulary scene graph generation.
\newblock In \emph{Proceedings of the AAAI Conference on Artificial Intelligence (AAAI)}, volume~39, pages 5576--5584, 2025.

\bibitem[Wang et~al.(2024{\natexlab{a}})Wang, Wu, Yang, and Luo]{Wang2024VidVRD}
Yongqi Wang, Xinxiao Wu, Shuo Yang, and Jiebo Luo.
\newblock End-to-end open-vocabulary video visual relationship detection using multi-modal prompting.
\newblock \emph{arXiv preprint arXiv:2409.12499}, 2024{\natexlab{a}}.

\bibitem[Fang et~al.(2019)Fang, Yan, Qiao, Xue, Wang, and Li]{fang2019dada}
Jianwu Fang, Dingxin Yan, Jiahuan Qiao, Jianru Xue, He~Wang, and Sen Li.
\newblock Dada-2000: Can driving accident be predicted by driver attentionƒ analyzed by a benchmark.
\newblock In \emph{2019 IEEE Intelligent Transportation Systems Conference (ITSC)}, pages 4303--4309. IEEE, 2019.

\bibitem[Wang et~al.(2024{\natexlab{b}})Wang, Kim, Wenxuan, Xie, Ge, Chen, Li, and Luo]{wang2024deepaccident}
Tianqi Wang, Sukmin Kim, Ji~Wenxuan, Enze Xie, Chongjian Ge, Junsong Chen, Zhenguo Li, and Ping Luo.
\newblock Deepaccident: A motion and accident prediction benchmark for v2x autonomous driving.
\newblock In \emph{Proceedings of the AAAI Conference on Artificial Intelligence}, volume~38, pages 5599--5606, 2024{\natexlab{b}}.

\bibitem[Bao et~al.(2020)Bao, Yu, and Kong]{bao2020uncertainty}
Wentao Bao, Qi~Yu, and Yu~Kong.
\newblock Uncertainty-based traffic accident anticipation with spatio-temporal relational learning.
\newblock In \emph{Proceedings of the 28th ACM International Conference on Multimedia}, pages 2682--2690, 2020.

\bibitem[Jocher et~al.(2023)]{ultralytics2023yolov8}
Glenn Jocher et~al.
\newblock Yolov8: Open-source object detection model.
\newblock \url{https://github.com/ultralytics/ultralytics}, 2023.
\newblock Accessed: 2025-05-10.

\bibitem[Aharon et~al.(2022)Aharon, Orfaig, and Bobrovsky]{aharon2022bot}
Nir Aharon, Roy Orfaig, and Ben-Zion Bobrovsky.
\newblock Bot-sort: Robust associations multi-pedestrian tracking.
\newblock \emph{arXiv preprint arXiv:2206.14651}, 2022.

\bibitem[Ren et~al.(2016)Ren, He, Girshick, and Sun]{ren2016faster}
Shaoqing Ren, Kaiming He, Ross Girshick, and Jian Sun.
\newblock Faster r-cnn: Towards real-time object detection with region proposal networks.
\newblock \emph{IEEE transactions on pattern analysis and machine intelligence}, 39\penalty0 (6):\penalty0 1137--1149, 2016.

\bibitem[Mitra et~al.(2024)Mitra, Huang, Darrell, and Herzig]{mitra2024compositional}
Chancharik Mitra, Brandon Huang, Trevor Darrell, and Roei Herzig.
\newblock Compositional chain-of-thought prompting for large multimodal models.
\newblock In \emph{Proceedings of the IEEE/CVF Conference on Computer Vision and Pattern Recognition}, pages 14420--14431, 2024.

\bibitem[Li et~al.(2024{\natexlab{d}})Li, Lin, Peng, Nyandwi, Jiang, Ma, Khanuja, Krishna, Neubig, and Ramanan]{li2024naturalbench}
Baiqi Li, Zhiqiu Lin, Wenxuan Peng, Jean de~Dieu Nyandwi, Daniel Jiang, Zixian Ma, Simran Khanuja, Ranjay Krishna, Graham Neubig, and Deva Ramanan.
\newblock Naturalbench: Evaluating vision-language models on natural adversarial samples.
\newblock \emph{arXiv preprint arXiv:2410.14669}, 2024{\natexlab{d}}.

\bibitem[Liu et~al.(2023)Liu, Li, Wu, and Lee]{liu2023visual}
Haotian Liu, Chunyuan Li, Qingyang Wu, and Yong~Jae Lee.
\newblock Visual instruction tuning.
\newblock \emph{Advances in neural information processing systems}, 36:\penalty0 34892--34916, 2023.

\end{thebibliography}
}

\newpage
\appendix

\section{Dataset details}
\subsection{Original DRAMA dataset}
\label{appx:raw-data}
The DRAMA (Driving Risk Assessment Mechanism with A captioning module) dataset comprises 17,785 interactive driving scenarios captured over 91 hours of urban driving in Tokyo, Japan. Each scenario includes high-resolution video frames (1928×1280 at 30 Hz) recorded using a SEKONIX SF332X-10X camera. Accompanying each scenario are detailed annotations encompassing risk localization bounding boxes and free-form natural language captions that elucidate the perceived risks from the driver's perspective. 

The dataset emphasizes diverse risk factors such as lane changes, cut-ins, traffic congestion, and interactions with parked vehicles. Additionally, it features multi-level question-answer pairs (e.g., what, where, why) to support tasks like visual question answering and explainable risk reasoning. 

\subsection{Data Filtering Implementation}
\label{appx:data-curation}

To construct DRAMA-X, we filtered 17,785 raw frames down to 5,686 using a YOLOv8n-based object detector focused on pedestrians and cyclists. Detections were retained if they satisfied: (i) at least 50\% of the bounding box within frame bounds, (ii) minimum size thresholds of 8\% image height and 1\% width, and (iii) a maximum of 3 instances per class per frame. Cyclists were identified by pairing person and bicycle detections with IoU > 0.3 and appropriate vertical alignment (person above bicycle). Frames lacking valid detections after filtering were discarded. The following format was used for DRAMA-X annotations to save the filtered objects post filtering:
\begin{tcolorbox}[colback=gray!5!white, colframe=gray!75!black, before skip=1mm, after skip=1mm, boxsep=0.0cm, middle=0.0cm, top=0.1cm, bottom=0.1cm, title=DRAMA-X Data Sample Format]
\begin{verbatim}
{
    "sample_n": {
        "image_path": "<image_path_to_sample_n>",
        "video_path": "<video_path_to_sample_n>",
        "Risk": "Yes",
        "Pedestrians": {
            "1": {
                "Box": [
                    1085,
                    782,
                    1148,
                    935
                ],
                "Intent": [],
                "Position": "",
                "Description": ""
            },
            ...
        },
        "Cyclists": {},
        "suggested_action": "be aware or cautious 
        (of the important object, in case it might effect in 
        future but no direct influence)"
    },
    ...
}
\end{verbatim}
\end{tcolorbox}
We retain the "suggested\_action" and "Risk" fields from the original dataset.
\subsection{Annotation Pipeline parameters}
\label{appx:data-annotation}
\subsubsection{Object Tracking}
We use YOLOv8 \cite{ultralytics2023yolov8} + BoT-SORT \cite{aharon2022bot} for person tracking while cycle detection and tracking is handled using a Faster R-CNN model with a ResNet-50-FPN backbone \cite{ren2016faster} +  BoT-SORT. We combine person and cycle detections with an overlap of > 0.3 and maximum vertical offset of 160 pixels. These combined bounding boxes are input to the tracking algorithm.

\begin{figure}
\centering
\includegraphics[width=\linewidth]{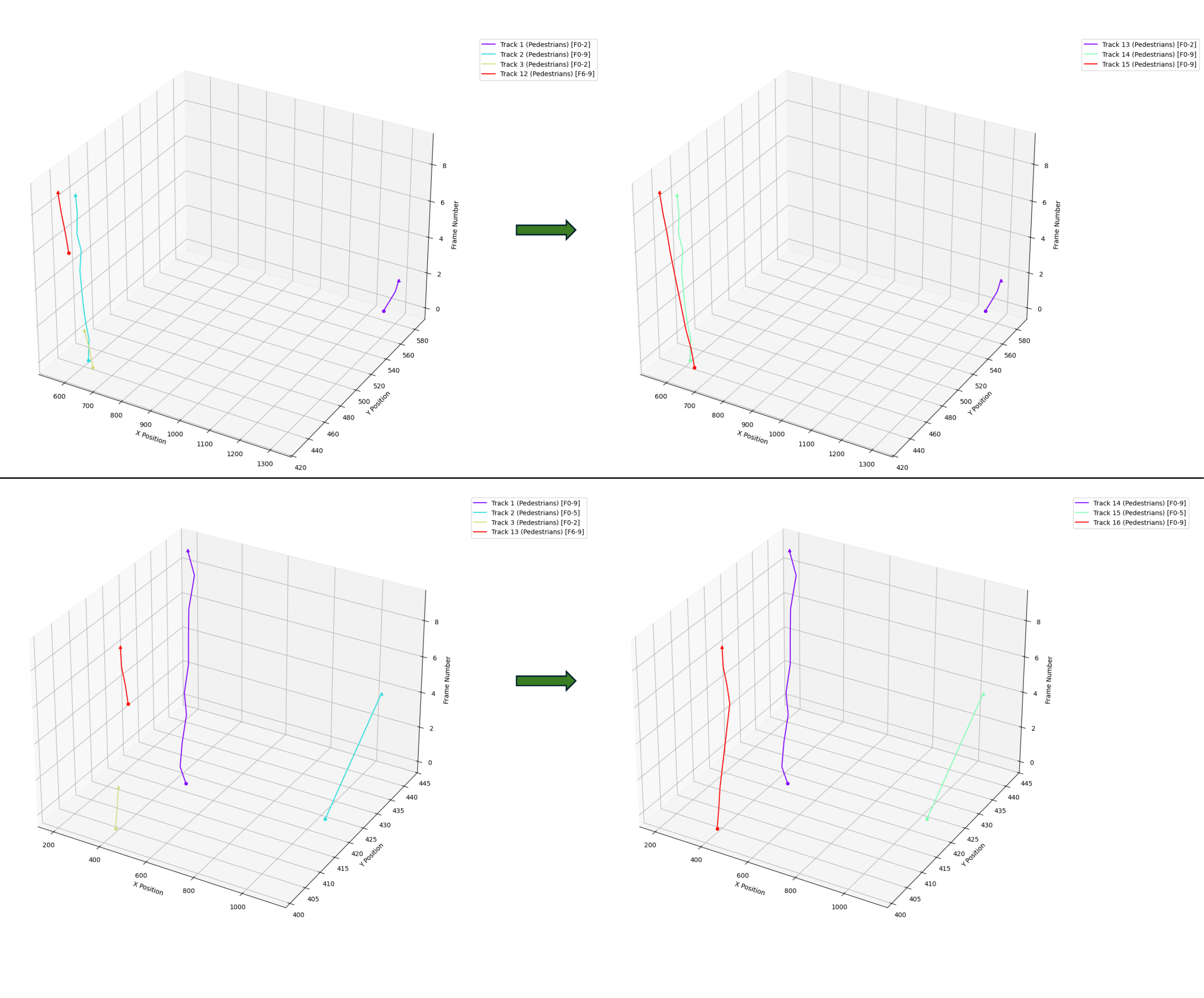}
\caption{Qualitative results of the track linking process. Left: fragmented tracks produced by BoT-SORT. Right: refined trajectories after applying our spatial-temporal linking algorithm.}
\label{fig:link_results}
\end{figure}

\paragraph{Track Linking} 
Despite the high tracking accuracy of BoT-SORT, fragmented object tracks still occur. To address this, we compute a linking score between tracks $i$ and $j$ as:
\begin{equation*}
S_{i,j} = \left(1 - \frac{d_{\text{spatial}}}{D_{\max}}\right) w_s + \left(1 - \frac{\Delta t}{T_{\max}}\right) w_t, \quad
\hat{S}{i,j} = S{i,j} \cdot (0.5 + 0.5\alpha)
\end{equation*}
where $\alpha \in [0, 1]$ is the appearance similarity, and $w_s$, $w_t$ are the spatial and temporal weights. $D_{\max}$ and $T_{\max}$ denote the maximum allowable spatial and temporal distances, adaptively determined by track length and frame gap. A link is accepted if $\hat{S}{i,j} > \theta_{\text{link}}$, with $\theta_{\text{link}} = 0.2$ for short gaps ($\Delta t \leq 3$) and $\theta_{\text{link}} = 0.3$ otherwise. Examples of object tracks before and after linking are shown in Figure~\ref{fig:link_results}.

\paragraph{Object Matching} 
For ground-truth boxes $\mathcal{B} = \{B_1, \dots, B_m\}$ with active tracked objects $\mathcal{T} = \{T_1, \dots, T_n\}$ we find the associated track using the cost matrix with: 
\begin{equation*}
\min_{\pi \in \text{Perm}(n, m)} \sum_{i=1}^{n} C_{i, \pi(i)} \quad \text{subject to } C_{i, \pi(i)} < 1 - \theta_{\text{iou}}
\end{equation*}
Where inverse IoU is $ 
C_{i,j} = 1 - \text{IoU}(T_i, B_j)$. We solve the optimal assignment using the Hungarian algorithm with $\theta_{\text{iou}} = 0.3$. When the solver fails (e.g., due to degenerate cost matrix), we fall back to a greedy assignment strategy that iteratively selects the lowest-cost valid pairs.

\subsection{Evaluation Tasks and Metrics Formulation}
\label{appx:evaluation}

\subsubsection{Object Detection (OD)}  
Fraction of ground‐truth objects correctly localized (IoU$\geq$0.5):
\[
\mathrm{Acc}_{\mathrm{OD}}
\;=\;
\frac{\bigl|\{\,(b_{\text{gt}},b_{\text{pred}})\mid \mathrm{IoU}(b_{\text{gt}},b_{\text{pred}})\ge0.5\}\bigr|}
{N_{\text{gt}}}
\]
where $N_{\text{gt}}$ is the number of ground‐truth boxes.

\subsubsection{Intent Prediction (IP)}  
We predict lateral and vertical intents over $\mathcal{I}$ classes.  Reported metrics:
\[
\mathrm{Acc}_{\mathrm{LIP}}
\;=\;
\frac{\#\{\hat{y}^{\mathrm{lat}}=y^{\mathrm{lat}}\}}{N},
\quad
\mathrm{Acc}_{\mathrm{VIP}}
\;=\;
\frac{\#\{\hat{y}^{\mathrm{ver}}=y^{\mathrm{ver}}\}}{N}
\]
\[
\quad
\mathrm{Acc}_{\mathrm{Combined}}
\;=\;
\frac{\#\{\hat{y}^{\mathrm{lat}}=y^{\mathrm{lat}}\;\wedge\;\hat{y}^{\mathrm{ver}}=y^{\mathrm{ver}}\}}{N}.
\]

\subsubsection{Risk Assessment (RA)}  
Binary classification of scene risk.  We report balanced accuracy and positive‐class F1:
\[
\mathrm{BA}
=\tfrac12\Bigl(\tfrac{\mathrm{TP}}{\mathrm{TP+FN}}+\tfrac{\mathrm{TN}}{\mathrm{TN+FP}}\Bigr),
\quad
\mathrm{F1}
=\frac{2\,\mathrm{TP}}{2\,\mathrm{TP}+\mathrm{FP}+\mathrm{FN}}.
\]

\subsubsection{Action Suggestion (AS)}  
Free‐form driving instructions evaluated with BERTScore$_{F1}$ to capture semantic alignment between generated and reference actions.

\subsection{DRAMA-X Example Annotations}

\begin{figure}[h]
\centering
\includegraphics[width=\linewidth]{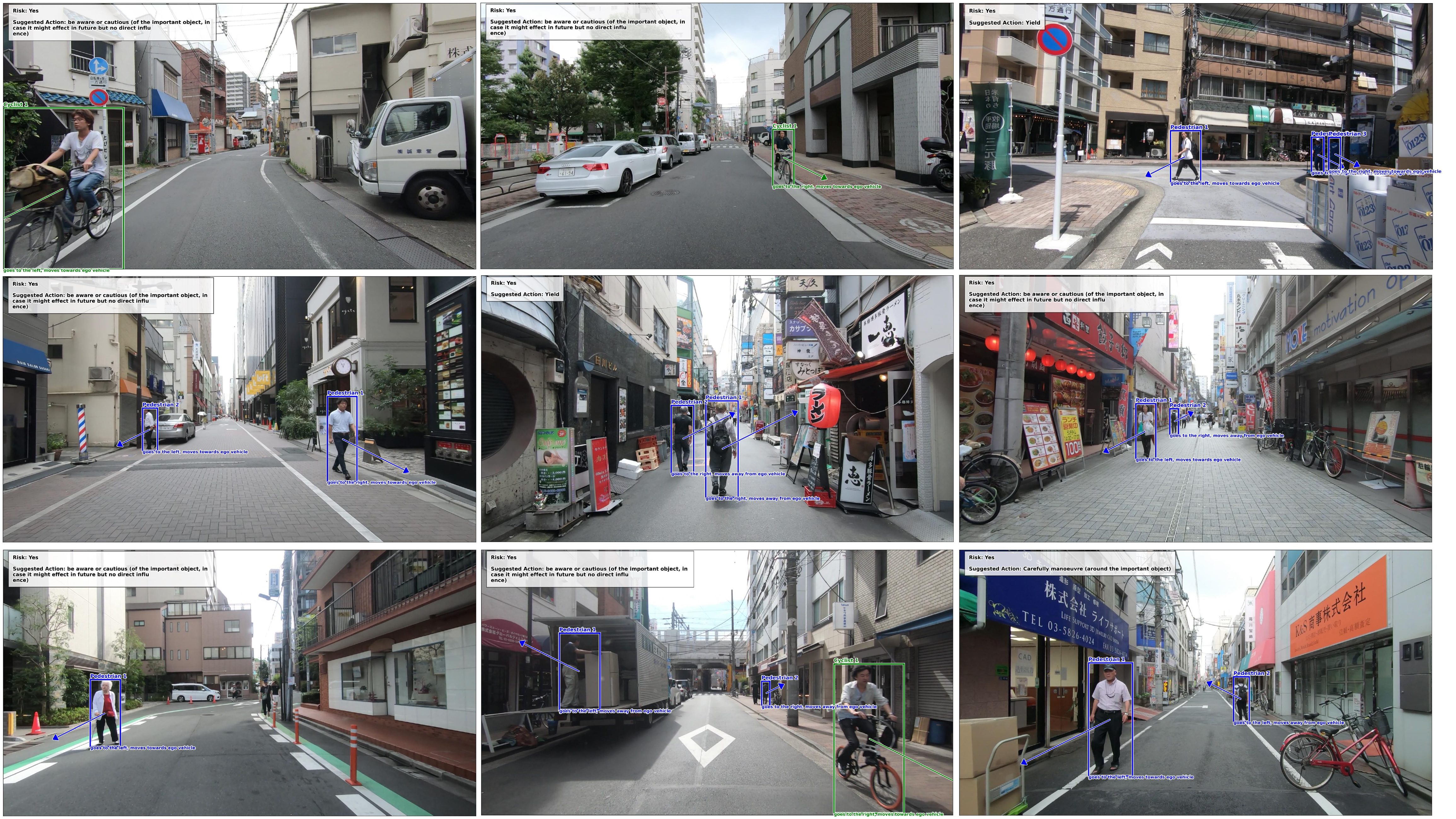}
\caption{Representative DRAMA-X annotations illustrating directional intent (lateral and vertical), associated risk scores (Yes/No), and recommended driving actions (e.g., yield, slow down).}

\label{fig:sample_annotations}
\end{figure}

Sample annotations demonstrating the intent vectors and associated risk scores are illustrated in Figure~\ref{fig:sample_annotations}.

\section{Implementation details for SGG-Intent Framework}
The framework includes a scene graph generation model and a VLM for intent prediction. In this work we use the same VLM for both tasks. The models used are detailed in Table~\ref{tab:models}.

\begin{table}[h]
\caption{Overview of vision language models evaluated on DRAMA-X.}
    \label{tab:models}
  \centering
  \small
  \begin{tabular}{lll}
    \toprule
    \textbf{Model Family} & \textbf{Model Name}               & \textbf{\#Parameters}    \\
    \midrule
    GPT-4                 & GPT-4o                           & Proprietary / undisclosed \\
    LLaVA                 & LLaVA-v1.6-mistral-7b            & 7B                       \\
    Qwen-VL               & Qwen2.5-VL-7B-Instruct           & 7B                       \\
    Molmo                 & Molmo-7B-D-0924                  & 7B                       \\
    \bottomrule
  \end{tabular}
  
\end{table}

 \subsection{Vision Language Model Prompts}
 Inspired from the zero-shot prompting stratgies to elicit scene graphs \cite{mitra2024compositional}, we design a prompt suitable for driving scenarios. This prompt is slightly tweaked to align the responses of different models. 
We utilize code from original model documentations available on \url{https://huggingface.co/} for all VLMs used. Following are the prompts used for scene graph generation, intent generation and raw one pass output generation respectively.
\begin{tcolorbox}[
  title=Scene Graph Generation Prompt,
  colback=gray!5!white,
  colframe=gray!75!black,
  boxsep=1mm,
  left=1mm, right=1mm,
  top=1mm, bottom=1mm,
  before skip=1mm, after skip=1mm,
  fontupper=\scriptsize\ttfamily,
  parbox=false
]
For the provided image, generate a scene graph in JSON format that includes the following, be concise and consider only important objects:

1. Objects in the frame. The special requirement is that you must include every pedestrian and cyclist separately and not group them as people or cyclists.

2. Object attributes inside object dictionary that are relevant to answering the question. Object attribute should include the state of the object eg. moving or static, description of the object such as color, orientation, etc.

3. Object bounding boxes. These should be with respect to the original image dimensions.

4. Object relationships between objects. This should be detailed of upto 4 words. 

Limit your response to only at most 5 most relevant objects in the scene.

An example structure would look like this:
\begin{verbatim}
{
  "Objects": {
    "name_of_object": {
      "attributes": [],
      "bounding_box": []
    }
  },
  "Relationships": {
    {
      "from": "name_of_object1",
      "to":   "name_of_object2",
      "relationship": "relationship between obj_1 and obj_2"
    }
  }
}
\end{verbatim}
Strictly output valid JSON only.

Scene Graph:
\end{tcolorbox}

\begin{tcolorbox}[
  title=Intent Generation Prompt,
  colback=gray!5!white,
  colframe=gray!75!black,
  boxsep=1mm,
  left=1mm, right=1mm,
  top=1mm, bottom=1mm,
  before skip=1mm, after skip=1mm,
  fontupper=\scriptsize\ttfamily,
  parbox=false
]
For the provided scene graph, image and question, generate an object-intent JSON which includes the following:

1. All objects from the scene graph.

2. Predicted intent for every object. Intent should be one of these values:

    2.1 Lateral (Sideways) Intent Options (choose one):
   
      - ``goes to the left''
      
      - ``goes to the right''

    2.2 Vertical Intent Options (choose one):
    
      - ``moves away from ego vehicle''
      
      - ``moves towards ego vehicle''
      
      - ``stationary''

3. Reason for this prediction.

4. Bounding box of the object, these should be with respect to orginal image dimensions.

An example structure (dictionary, not a list):
\begin{verbatim}
{
  "name_of_object": {
    "Intent": ["predicted lateral intent", "predicted vertical intent"],
    "Reason": "reason for this prediction",
    "Bounding_box": [x1, y1, x2, y2]
  }
}
\end{verbatim}
The Intent field list should ALWAYS have two values: one for lateral and one for vertical. Strictly output in valid JSON format. This JSON shoud NOT contain details from the scene graph such as the relationships or attributes, stick to the format mentioned above.
\end{tcolorbox}

\begin{tcolorbox}[
  title=Raw Output Generation Prompt,
  colback=gray!5!white,
  colframe=gray!75!black,
  boxsep=1mm,
  left=1mm, right=1mm,
  top=1mm, bottom=1mm,
  before skip=1mm, after skip=1mm,
  fontupper=\scriptsize\ttfamily,
  parbox=false
]
For the provided image and question, generate a object-intent JSON which includes the following:
1. AT MOST 5 objects from the scene including Pedestrians and Cylists.

2. Predicted intent for every object. Intent should be one of these values:

    2.1 Lateral (Sideways) Intent Options (has to be from these two options):
      - ``goes to the left''
      - ``goes to the right''

    2.2 Vertical Intent Options:
      -``moves away from ego vehicle''
      - ``moves towards ego vehicle''
      - ``stationary''

3. Risk score for this prediction (Yes or No). Risk is defined as a hazardous scenario that poses danger to the ego vehicle.

4. Bounding box of each object. these should be with respect to orginal image dimensions.

5. Suggested action given the scene and risk score.

An example structure would look like this:
\begin{verbatim}
{
  "Risk": "Yes/No",
  "Suggested_action": "suggested action for ego vehicle",
  "pedestrian": {
    "Intent": ["predicted lateral intent", "predicted vertical intent"],
    "Reason": "reason for this prediction",
    "Bounding_box": [x1, y1, x2, y2]
  },
  "car": {
    "Intent": ["predicted lateral intent", "predicted vertical intent"],
    "Reason": "reason for this prediction",
    "Bounding_box": [x1, y1, x2, y2]
  }
  ... (for all objects and NOT a list)
}
\end{verbatim}
The Intent field list should ALWAYS have two values: one for lateral and one for vertical. Strictly output in valid JSON format.
\end{tcolorbox}

\section{Broader Impact and Ethics Statement}

\subsection{Broader Impact Statement}
DRAMA-X introduces a structured benchmark for evaluating fine-grained intent prediction and risk reasoning in urban driving scenarios, with a focus on vulnerable road users such as pedestrians and cyclists. By addressing safety-critical behavior prediction and integrating compositional reasoning frameworks, this work supports the development of more interpretable and context-aware autonomous driving systems. The benchmark encourages research in areas where current vision-language models (VLMs) underperform, ultimately contributing to safer and more reliable autonomous vehicle deployments in complex, high-risk environments.

\subsection{Ethics Statement}
DRAMA-X is built upon the publicly available DRAMA dataset, which already includes measures such as anonymization of faces and vehicle license plates. Our annotation pipeline is fully automated and does not involve human subjects or crowdsourcing. All annotations are derived from visual data with safeguards to prevent privacy violations. While the benchmark promotes safety-focused AI development, we recognize the limitations of automated annotations and advocate for transparent reporting and human oversight in real-world applications. We emphasize the importance of inclusive testing across diverse demographics and scenarios to mitigate bias and unintended harm.

\end{document}